\documentclass[runningheads]{llncs}
\usepackage[T1]{fontenc}
\usepackage{graphicx}
\usepackage{tcolorbox}

\usepackage{times}
\usepackage{soul}
\usepackage{url}
\usepackage[utf8]{inputenc}
\usepackage[small]{caption}
\usepackage{amsmath}
\usepackage{amssymb}
\usepackage{algorithm}
\usepackage{booktabs}
\usepackage{tikz}
\usetikzlibrary{arrows}
\usetikzlibrary{shapes}
\usepackage{multirow}
\usepackage{pgfplots}
\usepackage{subcaption}

\newcommand{\subplotsscale}{0.55}
\newcommand{\subplotsscaleR}{0.6}
\DeclareMathOperator*{\argmax}{arg\,max}
\newcommand{\system}{{\sc GraLan}}
\usepackage{wrapfig}

\begin{document}
\title{The Graph Language: How Knowledge Graphs Speak to Large Language Models\thanks{This work was supported by STROKE 5.0. Progetto, sviluppo e sperimentazione di una piattaforma tecnologica di servizi di
intelligenza artificiale a supporto della gestione clinica integrata di eventi acuti di ictus. CUP:B29J23000430005.}}
\titlerunning{The Graph Language}
%
\author{Giuseppe Pirrò\orcidID{000-0002-7499-5798}}
\authorrunning{G. Pirrò}
%
\institute{DeMaCS, University of Calabria\\
\email{giuseppe.pirro@unical.it}}

\maketitle              

\begin{abstract}
%
%
Large Language Models (LLMs) excel at reasoning but benefit from grounding provided by Knowledge Graphs (KGs). However, integrating these paradigms is challenging. We introduce \system, which enables KGs to "speak" directly in the LLM's semantic space through relational tokens that preserve graph structure. \system's trainable \textit{language mediator} generates structured tokens for any frozen LLM, creating a foundation for knowledge-intensive applications. We demonstrate its effectiveness in question-answering by re-framing the task as entity classification over question-focused subgraphs. Experiments show that \system\ significantly outperforms existing methods, particularly on complex multi-hop reasoning tasks, establishing a new paradigm for KG-LLM integration that maintains structural fidelity while leveraging LLMs' reasoning capabilities.
\end{abstract}

\section{Introduction}
\label{sec:introduction}

Knowledge graphs (KGs) have emerged as a powerful formalism for organizing structured information across numerous domains, including large-scale encyclopedic resources (e.g., DBpedia, Wikidata) and specialized scientific databases (e.g., drug-gene interactions) \cite{wang2017knowledge}. Their graph-based representations encode entities as nodes and relations as edges, enabling the discovery of rich relational patterns and the execution of symbolic (e.g., SPARQL) queries. Meanwhile, the rise of large language models (LLMs)---such as the GPT-family \cite{brown2020language}, BERT \cite{devlin2019bert}, and T5~\cite{chen2023a} ---has yielded impressive progress in natural language processing, with applications ranging from open-domain question answering to dialogue systems \cite{zhao2023survey}. The question we address in this paper is: \textit{can we enable KGs to speak the LLM language?} Achieving this would allow us to combine the structured, precise, and interconnected nature of KGs with the versatile reasoning and generative capabilities of LLMs. This integration is essential to enhance LLM performance by reducing hallucinations and enabling complex reasoning tasks.

\noindent
\textbf{Research challenges.} While KGs excel at capturing structured, factual relationships, and LLMs demonstrate remarkable linguistic understanding, their fundamentally different knowledge representations create a semantic gap that hinders effective integration. Current approaches either compromise the rich structural information in KGs through flattening or serialization~\cite{wang2021kepler,baek2023knowledge}, or require expensive fine-tuning that risks catastrophic forgetting~\cite{sun2021,yu2022,yasunaga2022}. Parameter-efficient methods~\cite{FatemiHP24,chai2023graphllm} and iterative reasoning approaches~\cite{SunXTW0GNSG24} have made progress but still struggle with complex multi-hop reasoning tasks over heterogeneous graphs. {Bridging} structured KGs and powerful but text-oriented LLMs remains a nontrivial task \cite{sung2021can}.

\noindent
\textbf{Our Proposal.} We introduce \system\ (The Graph Language), a framework that bridges the gap between KGs and LLMs through semantic alignment rather than structural transformation. \system\ treats a frozen LLM as a universal reasoning engine and allows KGs to "speak" directly in the LLM's semantic space through \textit{relational tokens} - learned representations that encode the semantics of node, relation, and task-specific information. These tokens are composed into structured LLM input sequences that integrate KG-derived embeddings with task-specific context. The LLM processes these sequences to generate hidden states that function as neural workspaces where KG information and reasoning processes seamlessly merge. These distributed representations simultaneously encode both factual knowledge and logical relationships, enabling the model to generate responses that harmoniously blend structured information with task-specific reasoning within a unified computational architecture.
\system\ reimagines the KG-LLM integration problem as semantic alignment rather than structural transformation. Instead of forcing KGs to conform to LLM input formats as in prior work~\cite{tian2024graph,zhang2024question}, \system\ teaches KGs to communicate using a \textit{language} that preserves their structural integrity while remaining fluently interpretable by frozen LLMs. Unlike methods that use fixed graph extraction~\cite{he2024g} or rely on prompt engineering~\cite{tang2024higpt}, our approach enables zero-loss knowledge transfer between symbolic and neural representations, combining the precision of graph-structured reasoning with the flexible inference capabilities of language models.
By establishing a bidirectional communication channel between KGs and LLMs via the \textit{language mediator}, \system\ creates opportunities for structured knowledge extraction, fact verification, multi-modal reasoning, and continuous knowledge integration—all while maintaining computational efficiency through its modular, parameter-efficient design. These capabilities address the hallucination challenges~\cite{zhao2023,ji2023,bang2023} and factual inconsistencies that plague current LLMs when reasoning over structured knowledge. This enriched representation supports diverse downstream applications while preserving the LLM's general-purpose language capabilities and ensuring computational efficiency.

\subsection{Application to Question Answering} 
\begin{figure*}[!h]
    \centering
    \includegraphics[width=\textwidth]{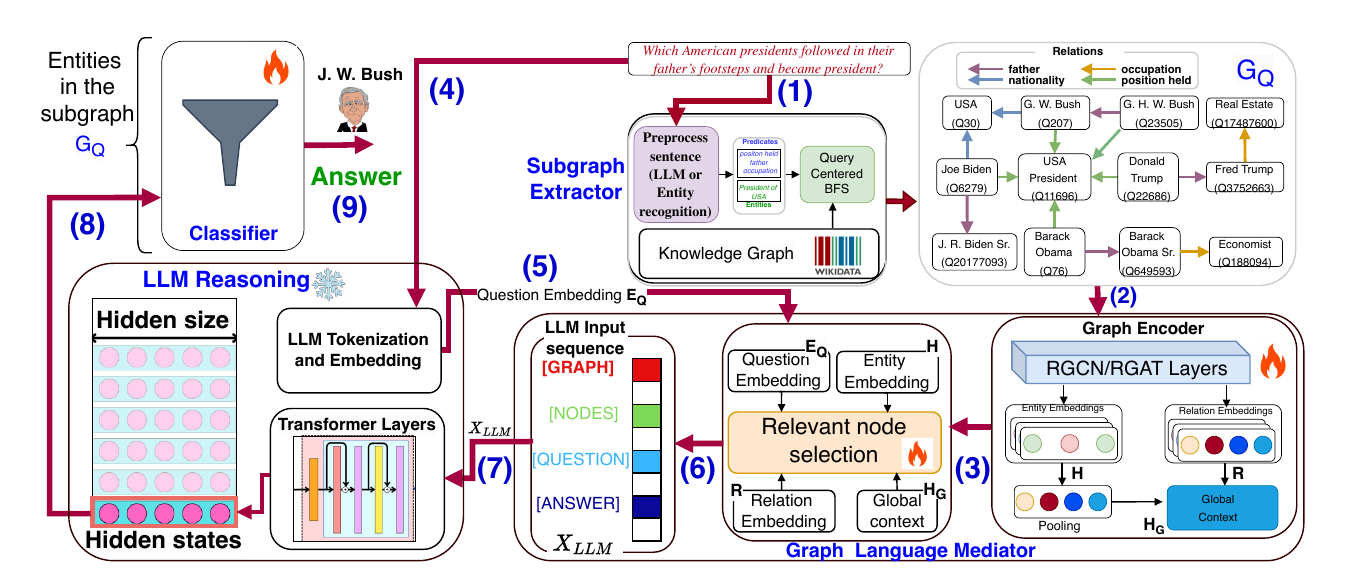}
    \vspace{-.8cm}
    \caption{\system's end-to-end architecture for graph-language reasoning. Given a natural language question (1), \system\  extracts a question-relevant subgraph (2) and processes it through RGCN/RGAT layers to obtain node (H), relation (R) and global context (H$_G$) embeddings (3). The question and graph representations are aligned in a shared semantic space through the Graph Language Mediator (4-6), which identifies the most related nodes to the question and creates a structured input sequence with special tokens [GRAPH], [NODES], [QUESTION], and [ANSWER] (during training). This sequence is processed by a frozen LLM (7) whose hidden states are used by a classifier (8) to predict the final answer, enabling reasoning without fine-tuning.}
    \label{fig:architecture}
    \vspace{-.4cm}
\end{figure*}

We show \system\ in the task of natural language question answering (QA).  The approach is outlined in Fig. \ref{fig:architecture}. Consider the following question from the GraptextQA dataset~\cite{shen2023graphextqa} using Wikidata as underlying KG and Flat-T5~\cite{raffel2020exploring} as LLM: 
\textit{“Which American presidents followed in their father’s footsteps and became president?”}. \system\ operates through three key components:

\noindent
(1) \underline{Question-Focused Subgraph:} rather than considering the entire KG, our framework first processes the natural language question to identify {relevant} seed entities and relations. This can be done via an LLM (with a carefully crafted prompt) or entity recognition. 
In this case (using ChatGPT) identified relevant entities are \textit{USA President} while relevant relations are \textit{father} and \textit{position held}. Then, \system\ uses a novel \textit{question-focused} breadth-first search algorithm on the underlying KG, which starting from the seed entities, \textit{USA President} in this case, \textit{biases} the traversal toward relations most related to \textit{father} and \textit{position held}. Part of the extracted subgraph is shown in Fig. \ref{fig:architecture}. There are several USA presidents along with their parents and the \textit{occupation} or \textit{position held}. We observe that \textit{occupation} is included in the subgraph because its related with the seed relation \textit{position held}. Less-relevant relations like \textit{part-of} or \textit{base salary}, that would have been considered by a traditional BFS, have been discarded thus reducing noise in the subsequent phases. 
%
\vspace{-.4cm}

\noindent
(2) \underline{Alignment:} the \textit{Graph Language Mediator} is responsible for transforming the structured information from the extracted subgraph into relational tokens that can be effectively utilized by the LLM.  First, a \textit{graph encoder} uses a GNN-architecture (e.g., RGCN) to embed entities (e.g., \textit{George H. W. Bush}) and their relations (e.g., \textit{father}, \textit{position held}) thus capturing their semantics in the subgraph. 
Besides entities and predicate embeddings, the encoder computes a \textit{global graph representation and a relation context embedding}. These embeddings encapsulate the overall structure of the subgraph and the specific relations \textit{pertinent to the query}. Through a cross-attention mechanism, entities dynamically align with the question's semantics (e.g., \textit{B. Obama}, \textit{J. W. Bush}, and \textit{D. J. Trump} nodes {attend} strongly to \textit{USA President} than \textit{Economist}) all in the LLM's native embedding space.
Then, the mediator assembles a specialized input sequence \( X_{LLM} \) by concatenating delimiter embeddings, a bird's-eye view of the entire subgraph (capturing global context around \textit{USA President}), embeddings of the most relevant nodes (e.g., \textit{J.W. Bush}), and the question representation itself.  

\noindent
(3) \underline{LLM reasoning:} \system\ treats the LLM as a general reasoning engine by keeping its parameters frozen. The frozen LLM processes the input sequence $\mathbf{X}_{\text{LLM}}$ and generates \textit{hidden states} for each token. The hidden state corresponding to the final token (denoted by \text{[ANSWER]}) encapsulates the necessary contextual information. In \system, the QA task is recast as \emph{predicting which entity in the subgraph} correctly answers the question. The LLM is guided to reason over our structured input and select \textit{J.W. Bush} as the answer directly from the subgraph entities through a classification layer. This formulation provides direct alignment between graph structure and LLM reasoning while remaining scalable to large KGs. By predicting entities rather than generating text, \system\ bypasses vocabulary constraints while maintaining the LLM's powerful reasoning capabilities.

\subsection{Contributions and outline}
Our approach advances the state of the art through six key innovations: 
(1) \textit{Preserving Structural Integrity} by directly \textit{encoding} question-relevant subgraphs with GNNs rather than flattening them into text; 
(2) \textit{Parameter-Efficient Integration} by keeping the LLM frozen, reducing computational costs while preventing catastrophic forgetting; 
(3) \textit{Reconceptualizing QA as Entity Classification} in the LLM's latent space, \textit{bypassing} token-vocabulary constraints and enabling linear scaling with KG size; 
(4) \textit{Unified Reasoning} through a relation-aware cross-attention mechanism that bridges symbolic KG structure and neural LLM reasoning; 
(5) \textit{Enhanced Scalability and Interpretability} via relatedness-based prioritization that efficiently scales to large KGs while ensuring explainable reasoning paths; and 
(6) A novel \textit{Graph Language Framework} where the Graph Language Mediator transforms graph structures into specialized relational tokens that serve as a semantic bridge between KGs and LLMs. 

\system\ offers flexibility, extending beyond question answering to support diverse applications including recommendation systems, knowledge discovery, and multi-modal reasoning without requiring architectural modifications to either system.

\smallskip
\noindent
The remainder of the paper is organized as follows. We review related work (\S\ref{sec:related-work}), then introduce the \system\ framework (\S\ref{sec:method}). Then, we report on an extensive experimental evaluation (\S\ref{sec:experiments}) and conclude (\S\ref{sec:conclusions}).
\section{Related Work}
\label{sec:related-work}

Integrating KGs with LLMs presents significant challenges due to their fundamentally different knowledge representation paradigms. We categorize existing approaches and highlight how \system\ advances the state of the art. Recent \textit{parameter-efficient} approaches with frozen LLMs have shown various limitations. {GraphToken}~\cite{FatemiHP24} serializes graphs into specialized token sequences but loses higher-order structural relationships, while {GraphLLM}~\cite{chai2023graphllm} creates ``graph-enhanced prefixes'' that capture only local neighborhood features. Both approaches overlook the inherent heterogeneity of real-world knowledge graphs (KGs). Task-specific approaches present additional constraints. {GNP}~\cite{tian2024graph} and {Q-KGR}~\cite{zhang2024question} remain limited to multiple-choice questions, while {HiGPT}~\cite{tang2024higpt} requires carefully crafted hierarchical prompts impractical for complex queries. Other approaches like {InstructGLM}~\cite{ye2023language} and {GraphGPT}~\cite{Tang00SSCY024} focus primarily on instruction tuning rather than semantic alignment. {LangGraph}\footnote{\url{https://github.com/langchain-ai/langgraph}} uses static, hand-crafted text templates to inject graph facts and treats graphs as input data processed through text workflows. In contrast, \system\ learns dense relational embeddings end-to-end via back-propagation from the QA loss, dynamically adapting to each query. \system\ treats graphs as first-class semantic objects communicating directly in the LLM's representation space, where learned tokens outperform fixed templates.

\noindent
While {G-Retriever}~\cite{he2024g} shares similarities with \system, it differs fundamentally in architectural philosophy. G-Retriever operates as a retrieval-augmented framework using Prize-Collecting Steiner Tree optimization and employs a dual-pathway approach combining textual serialization with Graph Attention Networks. In contrast, \system\ achieves true neural-symbolic integration through its Graph Language Mediator, establishing fundamental semantic alignment between graph structures and language understanding. \system's relation-aware tokens create a cohesive shared semantic space where graph structures and language coexist with minimal transformation loss, while its hybrid cross-attention classification mechanism operates directly in the LLM's latent space. Unlike \system's modular approach, {JAKET}~\cite{yu2022} requires complete joint pre-training of both KG and language components, increasing computational costs and complexity. Its bidirectional architecture fundamentally differs from \system's Graph Language Mediator, which preserves the LLM's capabilities while creating a specialized semantic alignment layer. While JAKET focuses on adapting to unseen KGs through extensive pre-training, \system\ offers greater versatility through its transparent task encoding mechanism without resource-intensive LLM re-training.

\noindent
\underline{\system's Key Advantages.} \system's advantages derive from a synergistic combination of innovations. Its parameter-efficient design keeps the LLM frozen, preventing catastrophic forgetting while reducing computational overhead. By reconceptualizing QA as entity classification in the LLM's latent space, \system\ transcends token-vocabulary limitations and scales linearly with KG size.  The Graph Language Mediator introduces three fundamental innovations that enable true neural-symbolic integration. First, unlike RAG approaches that flatten graphs into text, \system\ creates relational tokens that preserve graph structure while operating in the LLM's native semantic space, maintaining relational patterns through direct GNN encoding. Second, the mechanism enables bidirectional neural-symbolic communication through relation-aware cross-attention, going beyond simple concatenation or retrieval augmentation. Third, unlike G-Retriever's dual-pathway serialization or ToG's iterative exploration, \system\ maintains complete structural fidelity through direct graph-language integration without intermediate transformations. Beyond question answering, this approach serves as a universal translator between structured knowledge and language understanding, maintaining a consistent interface between KGs and LLMs across different domains while preserving task-specific semantics. This represents a paradigm shift from {adaptation-based} to {alignment-based} KG-LLM integration, fundamentally changing how knowledge graphs and language models interact and communicate.

\section{The Graph Language Framework}
\label{sec:method}
Let $G$= $(V, E, R)$ be a knowledge graph, where \(V = \{v_1, \dots, v_{|V|}\}\) is the set of entities, \(R = \{r_1, \dots, r_{|R|}\}\) is the set of relation types, and \(E \subseteq V \times R \times V\) is the set of edges (triplets); $\mathcal{N}_i^r$ is the set of neighbors of node $i$ linked by the relation  $r$. Each triplet \((v_i, r, v_j) \in E\) indicates a directed relational link \(r \in R\) from entity \(v_i\) to \(v_j\). Despite the versatility of \system, in this paper we focus on a concrete use case: natural language question answering combining KGs and LLMs. The problem can be stated as follows:

\noindent
\begin{tcolorbox}[
    colback=gray!10,
    colframe=gray!20,
    arc=4mm,
    boxrule=0.5pt,
    left=6pt,
    right=6pt,
    top=6pt,
    bottom=6pt,
    boxsep=0pt,
    width=\columnwidth,
    fontupper=\normalsize
]
\textbf{Problem}: Given a natural language query \(Q\), a knowledge graph $G$= $(V, E, R)$, and a LLM, the task is to predict the correct target entity \(v^* \in V\) that answers the question. 
\vspace{-.2cm}
\end{tcolorbox}

Our framework combines the KG's relational structure with linguistic information from \(Q\), ensuring an effective utilization by the LLM.
\system\ includes three primary modules: (i) task-focused subgraph extraction (\S\ref{sec:subgraph}); (ii) graph language mediator (\S\ref{sec:graph-language-mediator}); and (iii) LLM reasoner (\S\ref{sec:llm-reasoning}).
\subsection{Task-Focused Subgraph Extractor}
\label{sec:subgraph}
Given a knowledge graph \(\mathcal{G} = (V, E, R)\) and a natural language question \(Q\), we propose a two-stage process for extracting a task-relevant subgraph. There are three main reasons to focus on a local subgraph when answering questions over KGs. First,  most entities and relations are irrelevant to a specific query. For instance, the question, "Who manages the department where Alice works?" only involves exploring Alice's department and its managerial relations. Second, LLMs and GNNs have input size constraints; subgraphs make the input manageable while improving interpretability. Third, focusing on a smaller subgraph improves efficiency and accuracy by reducing processing time and irrelevant information~\cite{jiang2023unikgqa}.

\noindent
(1) \underline{Question Analysis:} the first stage involves mapping the natural language question \(Q\) to seed entities and relevant relations. This is achieved through a function \(\text{Map}({Q}) = (E_{\text{seed}}, P_{\text{rel}})\). Here, \(E_{\text{seed}} \subseteq V\) represent the entities identified as most relevant to the question (e.g., through Named Entity Recognition or using a carefully prompted LLM), and \(P_{\text{rel}} \subseteq R\) are the relations most aligned with the question's relations. The alignment is part of a question-focused BFS.

\noindent
(2) \underline{Question-Focused Breadth-First Search:} the second stage extracts the relevant subgraph $\mathcal{G}_{sub}$=$(V_{sub},R_{sub},E_{sub})$ based on the identified seed entities and relations. The subgraph is computed as\footnote{The complete pseudocode is available from the \system\ website \url{github.com/giuseppepirro/gralan}}:
		
\[
\mathcal{G}_{\text{sub}} = \text{QuestionFocusedBFS}({G}, E_{\text{seed}}, P_{\text{rel}}, \theta)
\]
where \(\theta\) is a relevance threshold that determines how closely edges in the graph must align with the question's context.
The edge expansion priority during BFS is determined by a similarity function \textit{sim}:$R$ $\times$ $R$ $\rightarrow [0,1]$, which measures the alignment between the relation \(r\) in the edge \(e = (v_i, r, v_j)\) and the question's seed relations \(P_{\text{rel}}\). The similarity function can be defined using either co-occurrence statistics within the KG~\cite{Pirro12} or embeddings of the relations. The algorithm maintains a priority queue \(\mathcal{Q}\), prioritizing edges with higher similarity scores. This ensures that the algorithm explores only the most relevant paths, focusing the subgraph around the seed entities while emphasizing edges (relations) most pertinent to the question relations. Furthermore, if multiple relevant relations or seed entities are identified during the Question Analysis stage, multiple BFSs can be initiated simultaneously with a final merge phase.
%
\subsection{Graph Language Mediator}
\label{sec:graph-language-mediator}

The Graph Language Mediator enables KGs to communicate with LLMs through three components:

\noindent
\textbf{Graph Encoder.} Using RGCN~\cite{schlichtkrull2018modeling}, it encodes the subgraph structure by initializing node embeddings $\mathbf{x}_i \in \mathbb{R}^d$ and relation embeddings $\mathbf{r}_k \in \mathbb{R}^d$. Through $L$ message-passing layers, we compute:
\begin{equation}
   \mathbf{h}_i^{(l)} = W_0^{(l)}\mathbf{h}_i^{(l-1)} + \sum_{r \in {R}} \sum_{j: (j, r, i) \in \mathcal{E}_{\text{sub}}} \frac{1}{|\mathcal{N}_i^r|} W_r^{(l)} \mathbf{h}_j^{(l-1)}
\end{equation}
This produces node embeddings $\mathbf{H}$ and relation embeddings $\mathbf{R}$. The global subgraph embedding $\mathbf{z}_{\text{graph}}$ is obtained via attention-weighted pooling~\cite{LiuZWLDHLT23} over the  embeddings:
\begin{align}
\small
   \alpha_i &= \text{softmax}(W_p\mathbf{h}_i), \\
   \mathbf{z}_{\text{graph}} &= \sum_i \alpha_i\mathbf{h}_i
\end{align}
where $W_p \in \mathbb{R}^{1 \times d}$ is a learnable parameter matrix.

\noindent
\textbf{Question Encoding.} The natural language question ${Q}$ is tokenized and embedded using the LLM vocabulary:
\begin{equation}
   \mathbf{E}_{Q} = \text{LLM.embed}({Q}) \in \mathbb{R}^{M \times d_{\text{lm}}}
\end{equation}
where $M$ is the number of tokens in the question, and $d_{\text{lm}}$ is the embedding dimension of the LLM.

\noindent
\textbf{Semantic Alignment.} To enable graph-question interaction, we employ a sequential process:

\noindent
1.\textit{Relation-Aware Node Interaction:} For nodes $i, j$ connected by relation $r$, we first compute attention scores considering their relational context:

\begin{equation}
\mathbf{A}_{i,j} = \text{softmax}\left(\frac{\mathbf{H}_i\mathbf{R}_r\mathbf{H}^{\top}_j \mathbf{E}^{\top}_Q}{\sqrt{d_{lm}}}\right)
\end{equation}

\noindent
It's important to note that $\mathbf{A}_{i,j}$ in Equation (5) serves a different purpose than the aggregation mechanism in Equation (1). While Equation (1) performs message passing based on predefined graph structure using the fixed neighborhood set $\mathcal{N}^r_i$, the attention scores $\mathbf{A}_{i,j}$ capture semantic relevance between connected nodes in the context of the question. These attention scores are used to enhance node representations before the question space projection, but do not directly replace the structural aggregation in the initial RGCN layers.

\noindent
2. \textit{Question Space Projection}: The relation-enriched representations are then projected into the question's semantic space via cross-attention:
\begin{align}
\small      
\label{eq:question-space-projection}
   \mathbf{Q} &= [\mathbf{H} \parallel \mathbf{R}] W_Q, \\
   \mathbf{K} &= \mathbf{E}_{Q} W_K, \\
   \mathbf{V} &= \mathbf{E}_{Q} W_V, \\
   \mathbf{Z} &= \text{softmax}\left(\frac{\mathbf{Q} \mathbf{K}^\top}{\sqrt{d}}\right)\mathbf{V}
\end{align}

\noindent
where $W_x$ are matrices. Each node-relation pair becomes:
\begin{equation}
   \mathbf{h}_i^{\text{final}} = \text{LayerNorm}\left(\mathbf{h}_i + \text{FFN}_g([\mathbf{h}_i; \mathbf{R}_i; \mathbf{z}_i])\right)
\end{equation}

\noindent
where $\text{FFN}_g$ is a feed-forward network of $g$ layers.

\smallskip
\noindent
\textbf{LLM Input Construction.} Then, \system\ additionally selects top-$K$ nodes embeddings $\mathbf{z}_{\text{top-}K}$ based on relevance to the question embedding $\mathbf{E}_Q$ and assembles the final sequence with special delimiters ($\mathbf{e}_{[\cdot]}$):

\begin{equation}
\mathbf{X}_{\text{LLM}} = [\mathbf{e}_{[\text{GRAPH}]}; \mathbf{z}_{\text{graph}}; \mathbf{e}_{[\text{NODES}]}; \mathbf{z}_{\text{top-}K}; \mathbf{e}_{[\text{QUESTION}]}; \mathbf{E}_Q; \mathbf{e}_{[\text{ANSWER}]}]
\end{equation}

\noindent
Note that in eq.(~\ref{eq:question-space-projection}), $\mathbf{Q}$ refers to the query matrix in the attention mechanism obtained by projecting node and relation embeddings, rather than the original question embedding $\mathbf{E}_Q$. This distinction allows the model to compute attention between graph elements and the question's semantic representation.

\subsection{LLM-Based Reasoning}
\label{sec:llm-reasoning}
After obtaining the input sequence $\mathbf{X}_{\text{LLM}}$, \system\ leverages a LLM for the reasoning and answer prediction phase. 

\noindent
\textbf{LLM Processing.}
The LLM processes the input sequence $\mathbf{X}_{\text{LLM}}$ through its transformer layers. This sequence allows us to bypass the usual token embedding lookup and directly leverage our graph-aligned representations. The LLM generates contextual representations:

\begin{equation}
    \{\mathbf{h}_{\text{LLM}}^{(t)}\}_{t=1}^L = \text{LLM}(\mathbf{X}_{\text{LLM}})
\end{equation}

\noindent
where $\mathbf{h}_{\text{LLM}}^{(t)} \in \mathbb{R}^{d_{\text{LLM}}}$ represents the hidden state at position $t$. Importantly, since the LLM remains frozen, all adaptation happens through the upstream graph mediator.

\subsection{End-to-End Training}
\label{sec:prediction}

\noindent
To predict answer entities from LLM outputs, we adopt a scalable classification approach. We extract the final hidden state corresponding to the [ANSWER] token:
\begin{equation}
   \mathbf{h}_{\text{answer}} = \mathbf{h}_{\text{LLM}}^{(\text{[ANSWER]})} \in \mathbb{R}^{d_{\text{LLM}}}
\end{equation}

\noindent
This representation is mapped to probabilities over subgraph entities:
\begin{equation}
   P_{\text{cls}}(v \mid X) = \text{softmax}\left({W}_{\text{cls}}\mathbf{h}_{\text{answer}}\right)_{id(v)}
\end{equation}
where ${W}_{\text{cls}} \in \mathbb{R}^{|V| \times d_{\text{LLM}}}$ projects to the entity space. The predicted entity is selected as:
\begin{equation}
   v_{\text{pred}} = \text{argmax}_{v \in V} P_{\text{cls}}(v \mid X)
   \label{eq:entity_prediction}
\end{equation}

\noindent
\textbf{Multiple Answer Handling:} While our core formulation focuses on single-answer prediction, \system\ can be extended to support multiple answers. Instead of selecting a single entity with $\argmax$ in eq. (\ref{eq:entity_prediction}), we apply a threshold $\tau$ to the entity probabilities:
\begin{equation}
V_{\text{pred}} = \{v \mid v \in V_{sub}, P_{\text{cls}}(v|X) > \tau\}
\end{equation}

\noindent
The threshold $\tau$ can be optimized on the validation set or set dynamically based on the distribution of probabilities. We implement this functionality by modifying the classifier to handle multi-label classification with independent binary decisions for each entity rather than a single decision.

\noindent
\textbf{Training Process.} For each batch of subgraph-query pairs, we compute both classification and LLM losses:
\begin{align*}
   P_{\text{cls}} &= \text{softmax}({W}_{\text{cls}}\mathbf{h}_{\text{answer}}) \\
   P_{\text{LLM}} &= \text{LLM}(X_{\text{LLM}})_{\text{output}} \\
   \mathcal{L}_{\text{total}} &= \lambda_{cls}\mathcal{L}_{\text{cls}}(P_{\text{cls}}, v^*) + (1-\lambda_{cls}) \mathcal{L}_{\text{LLM}}(P_{\text{LLM}}, v^*) +\alpha \mathcal{L}_{reg}
\end{align*}
where $v^*$ is the ground truth answer and $\mathcal{L}_{reg} = \|{W}_{\text{cls}}\|_F^2 + \beta\text{KL}(P_{\text{cls}} \| P_{\text{LLM}})$ is the regularization loss, where the first term is standard L2 regularization on the classification weights. The LLM loss $\mathcal{L}_{\text{LLM}}(P_{\text{LLM}}, v^*)$ is computed as the standard cross-entropy loss between the LLM output probabilities and the ground truth answer, encouraging the LLM component to assign high probability to the correct entity $
\mathcal{L}_{\text{LLM}}(P_{\text{LLM}}, v^*) = -\log(P_{\text{LLM}}(v^*))$. The second term uses Kullback-Leibler (KL) divergence to measure the difference between classification probabilities $P_{\text{cls}}$ and LLM probabilities $P_{\text{LLM}}$. This encourages the classification layer's predictions to stay semantically aligned with the LLM's understanding while allowing for graph-specific specialization. The hyperparameter $\beta$ controls this alignment strength while $\alpha$ and $\lambda_{cls}$ balance between classification learning, LLM-guided learning, and regularization strength.

\noindent
\system's approach offers key advantages: (i) Scalability through decoupling from LLM vocabulary size; (ii) flexibility via independent training of the classification layer; (iii) interpretability through classification probabilities; (iv) synergy between graph structure and LLM reasoning through balanced training objectives.
\section{Experiments}
\label{sec:experiments}
Given the vast amount of literature in KG-based QA, we evaluate the effectiveness of \system\ along two main lines: Open (\S \ref{sec:multi-relational-qa}) and multiple-choice (\S \ref{sec:multiple-choice-qa}) question answering. Then, we discuss an extensive ablation study (\S \ref{sec:ablation}).

\noindent
\textbf{Datasets.} As for \textit{Open QA}, we considered PathQuestion (PQ) and PathQuestion-Large (PQL) \cite{zhou2018interpretable} built on Freebase \cite{bollacker2008freebase}. PQL presents a more challenging scenario with reduced training data and larger KGs. MetaQA \cite{zhang2018variational}, derived from WikiMovies \cite{miller2016key}, focuses on movie-domain queries of varying complexity. Each dataset organizes questions by the number of reasoning hops required, enabling systematic evaluation of multi-hop reasoning capabilities. To benchmark against ToG \cite{SunXTW0GNSG24}, the current state-of-the-art LLM-based approach, we additionally include ZEROshotRE \cite{petroni2020kilt} and ComplexWebQuestions \cite{talmor2018web}. For these additional datasets, we utilize Wikidata 5M \cite{wang2021kepler} as KG. 
 As for \textit{multiple-choice QA}, we considered four benchmarks, OBDA \cite{mihaylov2018openbookqa}, ARC \cite{clark2018arc}, RIDDLE \cite{lin2021riddlesense} and PIQA \cite{bisk2020piqa}, all utilizing ConceptNet \cite{SpeerCH17} as the underlying knowledge graph\footnote{Preprocessed from DRAGON~\cite{yasunaga2022}}. Moreover, as done in~\cite{zhang2024question} we use all answer nodes and question nodes to extract a single subgraph (re-grounded) as the knowledge source for the LLM. the datasets are summarized in Table \ref{tab:datasets-appendix}. Further details are available in the Supplemental Material.
\begin{table*}[!t]
\centering
\small
  \resizebox{\textwidth}{!}{
    \setlength{\tabcolsep}{4pt}
    \renewcommand{\arraystretch}{1.2}
\begin{tabular}{@{}lcccccccc@{}}
\toprule
\textbf{Datasets} & \textbf{Task} & \textbf{KG} & \multicolumn{3}{c}{\textbf{Underlying KG Stats}} & \multicolumn{3}{c}{\textbf{Question Sets}} \\ 
\cmidrule(r){4-6} \cmidrule(l){7-9}
& & & \#Entities & \#Relations & \#Triples & \#Train & \#Valid & \#Test \\ \midrule
PQ-2h & Open QA & Freebase & 1,056 & 13 & 1,211 & 1,526 & 190 & 192 \\
PQ-3h & Open QA & Freebase & 1,836 & 13 & 2,839 & 4,158 & 519 & 521 \\
PQL-2h & Open QA & Freebase & 5,034 & 363 & 4,247 & 1,276 & 159 & 159 \\
PQL-3h & Open QA & Freebase & 6,505 & 411 & 5,597 & 825 & 103 & 103 \\
MetaQA 1-hop & Open QA & WikiMovies & 43,234 & 9 & 134,741 & 96,106 & 9,992 & 9,947 \\
MetaQA 2-hop & Open QA & WikiMovies & 43,234 & 9 & 134,741 & 118,980 & 14,872 & 14,872 \\
MetaQA 3-hop & Open QA & WikiMovies & 43,234 & 9 & 134,741 & 114,196 & 14,274 & 14,274 \\ \midrule
ZEROshotRE & Open QA (LLM comparison) & Wikidata 5M & $\sim$4.8M & 822 & 20.6M & 147,909 & 3,724 & 4,966 \\
ComplexWebQuestions & Open QA (LLM comparison)& Wikidata 5M& $\sim$4.8M & 822 & 20.6M & 27,734& 3,480& 3,531\\ \midrule
OBQA   & Multiple-choice QA & ConceptNet & $\sim$2M & 17 &$\sim$2.4M & 4,957 & 500 & 500 \\
ARC  & Multiple-choice QA & ConceptNet & $\sim$2M & 17 & $\sim$2.4M& {2,590} & 299 & 1,172 \\
RIDDLE  & Multiple-choice QA & ConceptNet & $\sim$2M & 17 & $\sim$2.4M& 3,510 & 1,021 & 1,184 \\
PIQA  & Multiple-choice QA & ConceptNet & $\sim$2M & 17 & $\sim2.4$M & 16,113 & 1,838 & 3,084 \\
\bottomrule
\end{tabular}}
\caption{Dataset statistics across different question-answering tasks.}
\label{tab:datasets-appendix}
\end{table*}

\noindent
\textbf{Experimental setting.}
\system\ is implemented in PyTorch(Geometric)\footnote{Code is available from GitHub {\url{https://github.com/giuseppepirro/gralan/}}}. To establish relation relatedness, we precomputed a TF-IDF-based relation relatedness matrix \cite{Pirro12}.
All experiments were conducted on a MacStudio with 192GB shared memory and are the average of 10 runs where the relatively low standard deviation is omitted.

\noindent
\underline{Hyperparameters.}
For each task-specific dataset, we tune hyperparameters using a validation set. For model architecture, we set the relational token dimension to match the LLM's hidden size, use 2 GNN layers with 5 attention heads, and employ a dropout rate of 0.1. The node selection mechanism considers the top-5 most relevant nodes for each question. We use the AdamW optimizer with a learning rate of 3e-5, batch size of 32, and weight decay of 0.01. We train for 100 epochs. The graph encoder's hidden size is 200, edge dimension of 64, and uses batch normalization. Details about hyperparameter tuning are available in the Supplemental Material.

\noindent
\underline{LLMs used.} While we conducted extensive experiments with various LLMs like BERT and GPT-2 (as detailed in our ablation studies \S\ref{sec:ablation}), we focus our discussion on FLAN-T5-XXL as the LLM reasoning engine, as it consistently outperformed other language models in our preliminary studies.

\noindent
\underline{Dataset splits.} For both Open and Multiple-choice QA, we used the original splits.

\noindent
\underline{Metric.} Exact
match accuracy (Hits@1) is used following previous work~\cite{SunXTW0GNSG24,WangRCB24}.
\subsection{Results on Open Questions}
\label{sec:multi-relational-qa}
We mainly choose reasoning-based QA methods as baselines: MINERVA~\cite{das2018go}: uses RL to learn paths through knowledge graphs, IRN~\cite{zhou2018interpretable}: employs an interpretable reasoning framework with RL, SRN~\cite{qiu2020structured}: leverages structured reasoning networks with policy learning, TransferNet~\cite{shi2021transfernet}: uses iterative label transfer, NSM~\cite{he2021improving}: neural state machine approach, JAKET \cite{yu2022}, G-Retriever \cite{he2024g}, and QAGCN \cite{WangRCB24} currently representing the most recent approach we are aware of.
Table~\ref{tab:results-QA-1} presents the performance comparison between \system\ and existing state-of-the-art approaches across different multi-hop reasoning tasks. Several key observations emerge from these results. First, on the PathQuestion datasets, \system\ consistently outperforms existing approaches across all configurations. The performance gap becomes particularly evident in the more challenging PQL dataset, where \system\ achieves substantial improvements of 3.25\% and 15.7\% over the previous best results for 2-hop and 3-hop reasoning, respectively. Notably, while other models show significant performance degradation when moving from PQ to the more complex PQL, \system\ maintains robust performance, demonstrating its superior ability to handle larger knowledge graphs with reduced training data.
\begin{table}[h]
    \centering
    \small
    \caption{Results on open questions based on multi-hop reasoning.}
    \setlength{\tabcolsep}{4pt}
    \renewcommand{\arraystretch}{1.2}
    \begin{tabular}{lcccccccc}
        \toprule
        \multirow{2}{*}{\textbf{Model}} & \multicolumn{4}{c}{\textbf{PathQuestion}} & \multicolumn{3}{c}{\textbf{MetaQA}} \\
        \cmidrule(lr){2-5} \cmidrule(lr){6-8}
        & \textbf{PQ-2h} & \textbf{PQ-3h} & \textbf{PQL-2h} & \textbf{PQL-3h} & \textbf{1-hop} & \textbf{2-hop} & \textbf{3-hop} \\
        \midrule
        MINERVA    & 75.9  & 71.2  & 71.8  & 65.7  & 96.3  & 92.9  & 55.2 \\
        IRN      & 91.9  & 83.3  & 63.0  & 61.8  & 85.9  & 71.3  & 35.6 \\
        EmbedKGQA   & 90.1  & 86.2  & 79.2  & 61.2  & 97.5  & 98.8  & 94.8 \\
        SRN        & 96.3  & 89.2  & 78.6  & 77.5  & 97.0  & 95.1  & 75.2 \\
        TransferNet& 91.1  & 96.5  & 54.7  & 62.1  & 97.5  & \textbf{100.0} & \textbf{100.0} \\
        NSM         & 94.2  & 97.1  & 74.2  & 67.0  & 97.3  & 99.9  & 98.9 \\
        QAGCN       & 98.5  & 90.6  & 87.5  & 70.9  & 97.3  & 99.9  & 97.6 \\
        JAKET & 93.5  & 91.2  & 83.1  & 67.6  & 93.5  & 73.4  & 71.2\\
        G-Retriever & 96.8  & 97.1  & 83.1  & 83.3  & 96.8  & 96.2  & 99.4\\
        \midrule
        \system\ & \textbf{98.9} & \textbf{97.3} & \textbf{90.75} & \textbf{86.6} & \textbf{97.8} & \textbf{100.0} & \textbf{100.0} \\
        \bottomrule
    \end{tabular}
    \label{tab:results-QA-1}
\end{table}

On the MetaQA benchmark, \system\ achieves comparable or superior performance to state-of-the-art methods. Particularly noteworthy is its consistent performance across all hop lengths, matching TransferNet's perfect score on 2-hop queries and NSM's near-perfect performance on 3-hop queries. This consistency across varying reasoning depths demonstrates \system's robust multi-hop reasoning capabilities. A distinguishing feature of \system\ is its ability to maintain high performance across both datasets without task-specific architectural modifications. This generalization capability, combined with its strong performance on both simple and complex queries, suggests that \system's graph-language integration approach effectively captures both local and global graph structures while leveraging the language model's semantic understanding.

Moreover, when compared specifically to G-Retriever and JAKET, \system\ demonstrates clear advantages in both benchmarks. On PathQuestion, \system\ improves over G-Retriever by +2.1 points on PQ-2h (98.9\% vs. 96.8\%) and +7.65 points on PQL-2h (90.8\% vs. 83.1 \%), and outperforms JAKET by +5.4 points (93.5 \% → 98.9\%) and +7.65 points (83.1\% → 90\%) on the same splits. Moreover, \system\ edges out  G-Retriever on PQL-3h and surpasses JAKET by a substantial margin of 19 points (67.6\% → 86.6\%). On MetaQA, \system\ consistently beats G-Retriever—e.g.\ by +3.8 points on 2-hop (100.0\% vs. 96.2\%) and +0.6 points on 3-hop (100.0\% vs. 99.4\%)—and outperforms JAKET by over 26 points on 2-hop (73.4\% → 100.0\%) and nearly 29 points on 3-hop (71.2\% → 100.0\%). These comparisons underscore that our semantic alignment approach not only rivals but often exceeds the performance of both retrieval-based and joint-attention methods.


\subsubsection{Comparison with LLM-only}
We compared \system\ with Think-on-Graphs (ToG) taking the best results reported in~\cite{SunXTW0GNSG24}. ToG enhances LLM reasoning by integrating structured knowledge from KGs, enabling iterative exploration of reasoning paths. It treats the LLM as an agent that interactively explores related entities and relations within a KG. Our experiments demonstrate \system's effectiveness compared to state-of-the-art LLM-KG approaches.

\begin{wraptable}{r}{0.6\textwidth}
    \centering
    \small
    \setlength{\tabcolsep}{4pt} 
    \renewcommand{\arraystretch}{1.2} 
    \begin{tabular}{lcc}
        \toprule
        \textbf{Model} & \textbf{ZeroShotRE} & \textbf{ComplexWebQ} \\
        \midrule
        ToG w/GPT-4~\cite{SunXTW0GNSG24} & 88.3 & 67.6 \\
        \midrule
        \system\         & \textbf{89.45} & \textbf{70.23} \\
        \bottomrule
    \end{tabular}
    \caption{\system\ vs iterative LLM–KG approaches.}
    \label{tab:comparison-llm}
\end{wraptable}

\noindent
The table shows that \system\ with FLAN-T5-XXL outperforms ToG using GPT-4 across both evaluated datasets.  \system\ achieves improvements of +1.15\% on ZeroShotRE (89.45\% vs.\ 88.3\%) and +2.63\% on ComplexWebQuestions (70.23\% vs.\ 67.6\%). These gains are particularly noteworthy given that ToG utilizes GPT-4, which is generally considered more powerful than FLAN-T5-XXL. This suggests that our hybrid architecture's structured approach to knowledge integration is more effective than purely iterative exploration strategies.


\subsection{Results on Multiple-choice Questions}
\label{sec:multiple-choice-qa}
We compare \system\ with several state-of-the-art approaches including methods that combine KGs with LLMs: KG Flattening \cite{yasunaga2022}, which flattens graph nodes into a sequence via relevance score (REL) ranking; KAPING \cite{baek2023knowledge}, which injects important KG triples from one-hop (OH) and two-hop (TH) neighborhoods; and GNP \cite{tian2024graph}, which uses GNNs to model retrieved subgraphs as soft prompts to augment LLMs.
%
%
\begin{table}[!h]
   \centering
   \small
   \caption{Results on multiple-choice questions. Results marked with $^*$ are taken from Q-KGR~\cite{zhang2024question}, which also uses FLAN-T5-XXL as the underlying LLM.}
   \renewcommand{\arraystretch}{1.2}
   \begin{tabular}{lccccc}
       \toprule
       \textbf{Method} & \textbf{OBQA} & \textbf{Riddle} & \textbf{ARC} & \textbf{PIQA} & \textbf{Average} \\
       \midrule
       LLM-only$^*$          & 76.80 & 61.37 & 68.93 & 56.58 & 65.92 \\
       REL$^*$                & 72.80 & 53.53 & 66.78 & 56.80 & 62.48 \\
       KAPING TH$^*$          & 60.60 & 48.43 & 57.25 & 53.21 & 54.87 \\
       KAPING OH$^*$          & 60.00 & 47.65 & 56.65 & 51.69 & 54.00 \\
       Prompt Tuning$^*$      & 78.80 & 61.37 & 74.85 & 61.26 & 69.07 \\
       Full Fine-tuning$^*$   & 89.40 & 80.78 & 76.82 & 65.61 & 78.15 \\
       LoRA$^*$              & 88.60 & 74.90 & 78.54 & 65.61 & 76.91 \\
       LoRA+GNP$^*$           & 89.60 & 76.67 & 78.71 & 65.94 & 77.73 \\
       LoRA+Q-KGR$^*$          & \textbf{90.00} & {80.98} & {79.78} & \textbf{90.08} & {85.21} \\ \midrule
        \system\         & {89.87} & \textbf{81.03} & \textbf{80.56} & {89.65} & \textbf{85.26} \\
       \bottomrule
   \end{tabular}
   \label{tab:evaluation-mqa}
\end{table}

We also consider LLM adaptation approaches: Prompt Tuning \cite{lester2021power}, which treats embedded knowledge as soft prompts prefixed to text questions; LoRA \cite{hu2022}, which updates partial LLM parameters; and standard Full Fine-Tuning. Finally, we compare with Q-KGR \cite{zhang2024question}, the current state-of-the-art that extends GNP with parameter-efficient KG-LLM integration. 
Our experimental results shown in Table~\ref{tab:evaluation-mqa} demonstrate \system's effectiveness compared to existing approaches. The baselines LLM-only (65.92\% average) and REL (62.48\% average) show the limitations of pure language modeling or pure relational approaches. Similarly, KAPING variants (TH: 54.87\%, OH: 54.00\%) underperform, suggesting that their graph-to-text transformation strategy loses crucial structural information. Parameter-efficient tuning methods show improved performance. Prompt Tuning achieves 69.07\% accuracy, while LoRA reaches 76.91\%. Adding neural processing (LoRA+GNP: 77.73\%) provides modest gains, indicating the value of structural information. LoRA+Q-KGR, achieves strong results (85.21\% average) via graph-question alignment.

\system\ achieves state-of-the-art results with 85.26\% \textit{average accuracy}, offering several key improvements. \textit{First}, \system\ shows consistent strong performance across all datasets, with particularly notable gains on complex reasoning tasks. On ARC, which requires multi-hop scientific reasoning, \system\ achieves 80.56\% accuracy compared to the previous best of 79.78\%. On Riddle, which tests abstract conceptual understanding, \system's 81.03\% accuracy surpasses the previous 80.98\%. \textit{Second}, \system\ maintains competitive performance on simpler tasks like OBQA (89.87\% vs 90.00\% previous best) while significantly improving on challenging scenarios. This suggests that our relation-aware attention mechanism effectively captures both straightforward and complex reasoning patterns. Most notably, \system\ approaches or matches the best results on PIQA (89.65\% vs 90.08\%), despite PIQA's focus on physical commonsense reasoning—a traditional challenge for graph-based methods. This underlines \system's ability to combine structured knowledge with the LLM's reasoning.

\smallskip
\noindent
\underline{Efficiency Considerations.} While full fine-tuning achieves strong results (78.15\% average), it requires extensive computational resources and risks catastrophic forgetting. \system\ matches or exceeds these results while keeping the LLM frozen, offering a more practical solution for real-world applications. The performance gains over other parameter-efficient methods like LoRA (76.91\%) and LoRA+GNP (77.73\%) validate our approach of teaching the graph encoder to communicate directly in the LLM's semantic space. These results demonstrate that \system's novel approach to graph-language integration offers meaningful improvements over existing methods while maintaining efficiency. 

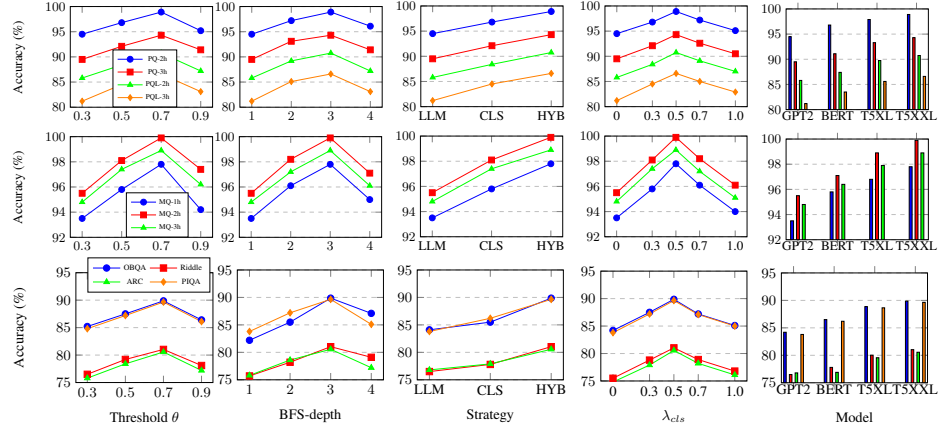
\begin{figure*}[!t]
\small
   \centering
   \begin{subfigure}[b]{0.19\textwidth}
       \centering
       \begin{tikzpicture}[scale=\subplotsscale]
       \begin{axis}[
           ylabel={Accuracy (\%)},
           ymin=80, ymax=100,
           legend style={font=\tiny, at={(0.75,0)}, anchor=south east},
           xtick={0.3, 0.5, 0.7, 0.9},
           ymajorgrids=true,
           grid style=dashed,
           width=5cm,
           height=4cm,
           legend entries={PQ-2h, PQ-3h, PQL-2h, PQL-3h},
       ]
       \addplot[color=blue,mark=*] coordinates {(0.3, 94.5) (0.5, 96.8) (0.7, 98.9) (0.9, 95.2)};
       \addplot[color=red,mark=square*] coordinates {(0.3, 89.5) (0.5, 92.1) (0.7, 94.3) (0.9, 91.4)};
       \addplot[color=green,mark=triangle*] coordinates {(0.3, 85.8) (0.5, 88.4) (0.7, 90.75) (0.9, 87.2)};
       \addplot[color=orange,mark=diamond*] coordinates {(0.3, 81.2) (0.5, 84.5) (0.7, 86.6) (0.9, 83.1)};
       \end{axis}
       \end{tikzpicture}
   \end{subfigure}
   \hspace{.15cm}
   \hfill
   \begin{subfigure}[b]{0.19\textwidth}
       \centering
       \begin{tikzpicture}[scale=\subplotsscale]
       \begin{axis}[
           ymin=80, ymax=100,
           legend style={font=\tiny, at={(0.95,0.15)}, anchor=south east},
           xtick={1, 2, 3, 4},
           ymajorgrids=true,
           grid style=dashed,
           width=5cm,
           height=4cm,
       ]
       \addplot[color=blue,mark=*] coordinates {(1, 94.5) (2, 97.2) (3, 98.9) (4, 96.1)};
       \addplot[color=red,mark=square*] coordinates {(1, 89.5) (2, 93.1) (3, 94.3) (4, 91.4)};
       \addplot[color=green,mark=triangle*] coordinates {(1, 85.8) (2, 89.2) (3, 90.75) (4, 87.2)};
       \addplot[color=orange,mark=diamond*] coordinates {(1, 81.2) (2, 85.1) (3, 86.6) (4, 83.1)};
       \end{axis}
       \end{tikzpicture}
   \end{subfigure}
   \hfill
   \begin{subfigure}[b]{0.19\textwidth}
       \centering
       \begin{tikzpicture}[scale=\subplotsscale]
       \begin{axis}[
           ymin=80, ymax=100,
           legend style={font=\tiny, at={(0.95,0.15)}, anchor=south east},
           symbolic x coords={LLM,CLS,HYB},
           xtick=data,
           ymajorgrids=true,
           grid style=dashed,
           width=5cm,
           height=4cm,
       ]
       \addplot[color=blue,mark=*] coordinates {(LLM, 94.5) (CLS, 96.8) (HYB, 98.9)};
       \addplot[color=red,mark=square*] coordinates {(LLM, 89.5) (CLS, 92.1) (HYB, 94.3)};
       \addplot[color=green,mark=triangle*] coordinates {(LLM, 85.8) (CLS, 88.4) (HYB, 90.75)};
       \addplot[color=orange,mark=diamond*] coordinates {(LLM, 81.2) (CLS, 84.5) (HYB, 86.6)};
       \end{axis}
       \end{tikzpicture}
   \end{subfigure}
   \hfill
   \begin{subfigure}[b]{0.19\textwidth}
       \centering
       \begin{tikzpicture}[scale=\subplotsscale]
       \begin{axis}[
           ymin=80, ymax=100,
           legend style={font=\tiny, at={(0.95,0.15)}, anchor=south east},
           xtick={0, 0.3, 0.5, 0.7, 1.0},
           xticklabels={0, 0.3, 0.5, 0.7, 1.0},
           ymajorgrids=true,
           grid style=dashed,
           width=5cm,
           height=4cm,
       ]
       \addplot[color=blue,mark=*] coordinates {(0, 94.5) (0.3, 96.8) (0.5, 98.9) (0.7, 97.2) (1.0, 95.1)};
       \addplot[color=red,mark=square*] coordinates {(0, 89.5) (0.3, 92.1) (0.5, 94.3) (0.7, 92.6) (1.0, 90.5)};
       \addplot[color=green,mark=triangle*] coordinates {(0, 85.8) (0.3, 88.4) (0.5, 90.75) (0.7, 89.1) (1.0, 87.0)};
       \addplot[color=orange,mark=diamond*] coordinates {(0, 81.2) (0.3, 84.5) (0.5, 86.6) (0.7, 85.0) (1.0, 82.9)};
       \end{axis}
       \end{tikzpicture}
   \end{subfigure}
   \hfill
   \begin{subfigure}[b]{0.19\textwidth}
       \centering
       \begin{tikzpicture}[scale=\subplotsscale]
       \begin{axis}[
           ymin=80, ymax=100,
           legend style={font=\tiny, at={(0.95,0.15)}, anchor=south east},
           symbolic x coords={GPT2,BERT,T5XL,T5XXL},
           xtick=data,
           ymajorgrids=true,
           grid style=dashed,
           width=5cm,
           height=4cm,
           ybar,
           bar width=1.6pt,
           x tick label style={rotate=0,anchor=center},
       ]
       \addplot[fill=blue] coordinates {(GPT2, 94.5) (BERT, 96.8) (T5XL, 97.9) (T5XXL, 98.9)};
       \addplot[fill=red] coordinates {(GPT2, 89.5) (BERT, 91.1) (T5XL, 93.3) (T5XXL, 94.3)};
       \addplot[fill=green] coordinates {(GPT2, 85.8) (BERT, 87.4) (T5XL, 89.75) (T5XXL, 90.75)};
       \addplot[fill=orange] coordinates {(GPT2, 81.2) (BERT, 83.5) (T5XL, 85.6) (T5XXL, 86.6)};
       \end{axis}
       \end{tikzpicture}
   \end{subfigure}
   \begin{subfigure}[b]{0.19\textwidth}
       \centering
       \begin{tikzpicture}[scale=\subplotsscale]
       \begin{axis}[
           ylabel={Accuracy (\%)},
           ymin=92, ymax=100,
           legend style={font=\tiny, at={(0.83,0)}, anchor=south east},
           xtick={0.3, 0.5, 0.7, 0.9},
           ymajorgrids=true,
           grid style=dashed,
           width=5cm,
           height=4cm,
           legend entries={MQ-1h, MQ-2h, MQ-3h},
       ]
       \addplot[color=blue,mark=*] coordinates {
           (0.3, 93.5) (0.5, 95.8) (0.7, 97.8) (0.9, 94.2)
       };
       \addplot[color=red,mark=square*] coordinates {
           (0.3, 95.5) (0.5, 98.1) (0.7, 99.9) (0.9, 97.4)
       };
       \addplot[color=green,mark=triangle*] coordinates {
           (0.3, 94.8) (0.5, 97.4) (0.7, 98.9) (0.9, 96.2)
       };
       \end{axis}
       \end{tikzpicture}
   \end{subfigure}
         \hspace{.14cm}
   \hfill
   \begin{subfigure}[b]{0.19\textwidth}
       \centering
       \begin{tikzpicture}[scale=\subplotsscale]
       \begin{axis}[
           ymin=92, ymax=100,
           legend style={font=\tiny, at={(0.95,0.15)}, anchor=south east},
           xtick={1, 2, 3, 4},
           ymajorgrids=true,
           grid style=dashed,
           width=5cm,
           height=4cm,
       ]
       \addplot[color=blue,mark=*] coordinates {
           (1, 93.5)  (2, 96.1)(3, 97.8) (4, 95.0)
       };
       \addplot[color=red,mark=square*] coordinates {
           (1, 95.5)  (2, 98.2)(3, 99.9) (4, 97.1)
       };
       \addplot[color=green,mark=triangle*] coordinates {
           (1, 94.8)  (2, 97.2) (3, 98.9)(4, 96.1)
       };
       \end{axis}
       \end{tikzpicture}
   \end{subfigure}
   \hfill
   \begin{subfigure}[b]{0.19\textwidth}
       \centering
       \begin{tikzpicture}[scale=\subplotsscale]
       \begin{axis}[
           ymin=92, ymax=100,
           legend style={font=\tiny, at={(0.95,0.15)}, anchor=south east},
           symbolic x coords={LLM,CLS,HYB},
           xtick=data,
           ymajorgrids=true,
           grid style=dashed,
           width=5cm,
           height=4cm,
       ]
       \addplot[color=blue,mark=*] coordinates {
           (LLM, 93.5) (CLS, 95.8) (HYB, 97.8)
       };
       \addplot[color=red,mark=square*] coordinates {
           (LLM, 95.5) (CLS, 98.1) (HYB, 99.9)
       };
       \addplot[color=green,mark=triangle*] coordinates {
           (LLM, 94.8) (CLS, 97.4) (HYB, 98.9)
       };
       \end{axis}
       \end{tikzpicture}
   \end{subfigure}
   \hfill
   \begin{subfigure}[b]{0.19\textwidth}
       \centering
       \begin{tikzpicture}[scale=\subplotsscale]
       \begin{axis}[
           ymin=92, ymax=100,
           legend style={font=\tiny, at={(0.95,0.15)}, anchor=south east},
           xtick={0, 0.3, 0.5, 0.7, 1.0},
           xticklabels={0, 0.3, 0.5, 0.7, 1.0},
           ymajorgrids=true,
           grid style=dashed,
           width=5cm,
           height=4cm,
       ]
       \addplot[color=blue,mark=*] coordinates {
           (0, 93.5) (0.3, 95.8) (0.5, 97.8) (0.7, 96.1) (1.0, 94.0)
       };
       \addplot[color=red,mark=square*] coordinates {
           (0, 95.5) (0.3, 98.1) (0.5, 99.9) (0.7, 98.2) (1.0, 96.1)
       };
       \addplot[color=green,mark=triangle*] coordinates {
           (0, 94.8) (0.3, 97.4) (0.5, 98.9) (0.7, 97.2) (1.0, 95.1)
       };
       \end{axis}
       \end{tikzpicture}
   \end{subfigure}
   \hfill
   \begin{subfigure}[b]{0.19\textwidth}
       \centering
       \begin{tikzpicture}[scale=\subplotsscale]
       \begin{axis}[
           ymin=92, ymax=100,
           legend style={font=\tiny, at={(0.95,0.15)}, anchor=south east},
           symbolic x coords={GPT2,BERT,T5XL,T5XXL},
           xtick=data,
           ymajorgrids=true,
           grid style=dashed,
           width=5cm,
           height=4cm,
           ybar,
           bar width=2pt,
           x tick label style={rotate=0,anchor=center},
       ]
       \addplot[fill=blue] coordinates {
           (GPT2, 93.5) (BERT, 95.8) (T5XL, 96.8) (T5XXL, 97.8)
       };
       \addplot[fill=red] coordinates {
           (GPT2, 95.5) (BERT, 97.1) (T5XL, 98.9) (T5XXL, 99.9)
       };
       \addplot[fill=green] coordinates {
           (GPT2, 94.8) (BERT, 96.4) (T5XL, 97.9) (T5XXL, 98.9)
       };
       \end{axis}
       \end{tikzpicture}
   \end{subfigure}
\begin{subfigure}[b]{0.19\textwidth}
    \centering
    \begin{tikzpicture}[scale=\subplotsscaleR]
    \begin{axis}[
        xlabel={Threshold $\theta$},
        ylabel={Accuracy (\%)},
        ymin=75, ymax=95,
        legend style={font=\tiny, at={(.98,0.82)}, anchor=south east,        legend columns=2},
        xtick={0.3, 0.5, 0.7, 0.9},
        ymajorgrids=true,
        grid style=dashed,
        width=4.6cm,
        height=4cm,
        legend entries={OBQA, Riddle, ARC, PIQA},
    ]
    \addplot[color=blue,mark=*] coordinates {
        (0.3, 85.2) (0.5, 87.5) (0.7, 89.87) (0.9, 86.4)
    };
    \addplot[color=red,mark=square*] coordinates {
        (0.3, 76.5) (0.5, 79.2) (0.7, 81.03) (0.9, 78.1)
    };
    \addplot[color=green,mark=triangle*] coordinates {
        (0.3, 75.8) (0.5, 78.4) (0.7, 80.56) (0.9, 77.2)
    };
    \addplot[color=orange,mark=diamond*] coordinates {
        (0.3, 84.8) (0.5, 87.2) (0.7, 89.65) (0.9, 86.1)
    };
    \end{axis}
    \end{tikzpicture}
\end{subfigure}
       \hspace{.18cm}
   \hfill
\begin{subfigure}[b]{0.19\textwidth}
    \centering
    \begin{tikzpicture}[scale=\subplotsscaleR]
    \begin{axis}[
        xlabel={BFS-depth},
        ymin=75, ymax=95,
        legend style={font=\tiny, at={(0.98,0.02)}, anchor=south east},
        xtick={1, 2, 3, 4},
        ymajorgrids=true,
        grid style=dashed,
        width=4.8cm,
        height=4cm,
    ]
    \addplot[color=blue,mark=*] coordinates {
        (1, 82.2) (2, 85.5)  (3, 89.87) (4, 87.1)
    };
    \addplot[color=red,mark=square*] coordinates {
        (1, 75.7) (2, 78.2) (3, 81.03)  (4, 79.1)
    };
    \addplot[color=green,mark=triangle*] coordinates {
        (1, 75.8) (2, 78.6) (3, 80.56)  (4, 77.2)
    };
    \addplot[color=orange,mark=diamond*] coordinates {
        (1, 83.8) (2, 87.2) (3, 89.65)  (4, 85.1)
    };
    \end{axis}
    \end{tikzpicture}
\end{subfigure}
   \hfill
\begin{subfigure}[b]{0.19\textwidth}
    \centering
    \begin{tikzpicture}[scale=\subplotsscaleR]
    \begin{axis}[
        xlabel={Strategy},
        ymin=75, ymax=95,
        legend style={font=\tiny, at={(0.98,0.02)}, anchor=south east},
        symbolic x coords={LLM,CLS,HYB},
        xtick=data,
        ymajorgrids=true,
        grid style=dashed,
        width=4.8cm,
        height=4cm,
    ]
    \addplot[color=blue,mark=*] coordinates {
        (LLM, 84.1) (CLS, 85.5) (HYB, 89.87)
    };
    \addplot[color=red,mark=square*] coordinates {
        (LLM, 76.5) (CLS, 77.8) (HYB, 81.03)
    };
    \addplot[color=green,mark=triangle*] coordinates {
        (LLM, 76.8) (CLS, 77.9) (HYB, 80.56)
    };
    \addplot[color=orange,mark=diamond*] coordinates {
        (LLM, 83.8) (CLS, 86.2) (HYB, 89.65)
    };
    \end{axis}
    \end{tikzpicture}
\end{subfigure}
   \hfill
\begin{subfigure}[b]{0.19\textwidth}
    \centering
    \begin{tikzpicture}[scale=\subplotsscaleR]
    \begin{axis}[
        xlabel={$\lambda_{cls}$},
        ymin=75, ymax=95,
        legend style={font=\tiny, at={(0.98,0.02)}, anchor=south east},
        xtick={0, 0.3, 0.5, 0.7, 1.0},
        xticklabels={0, 0.3, 0.5, 0.7, 1.0},
        ymajorgrids=true,
        grid style=dashed,
        width=4.8cm,
        height=4cm,
    ]
    \addplot[color=blue,mark=*] coordinates {
        (0, 84.2) (0.3, 87.5) (0.5, 89.87) (0.7, 87.2) (1.0, 85.1)
    };
    \addplot[color=red,mark=square*] coordinates {
        (0, 75.5) (0.3, 78.8) (0.5, 81.03) (0.7, 78.9) (1.0, 76.8)
    };
    \addplot[color=green,mark=triangle*] coordinates {
        (0, 74.8) (0.3, 77.9) (0.5, 80.56) (0.7, 78.2) (1.0, 76.1)
    };
    \addplot[color=orange,mark=diamond*] coordinates {
        (0, 83.8) (0.3, 87.2) (0.5, 89.65) (0.7, 87.1) (1.0, 85.0)
    };
    \end{axis}
    \end{tikzpicture}
\end{subfigure}
   \hfill
\begin{subfigure}[b]{0.19\textwidth}
       \centering
       \begin{tikzpicture}[scale=\subplotsscaleR]
       \begin{axis}[
           xlabel={Model},
           ymin=75, ymax=95,
           legend style={font=\tiny, at={(0.98,0.02)}, anchor=south east},
           symbolic x coords={GPT2,BERT,T5XL,T5XXL},
           xtick=data,
           ymajorgrids=true,
           grid style=dashed,
           width=4.8cm,
           height=4cm,
           ybar,
           bar width=1.6pt,
           x tick label style={rotate=0,anchor=center},
       ]
       \addplot[fill=blue] coordinates {
           (GPT2, 84.2) (BERT, 86.5) (T5XL, 88.87) (T5XXL, 89.87)
       };
       \addplot[fill=red] coordinates {
           (GPT2, 76.5) (BERT, 77.8) (T5XL, 80.03) (T5XXL, 81.03)
       };
       \addplot[fill=green] coordinates {
           (GPT2, 76.8) (BERT, 76.9) (T5XL, 79.56) (T5XXL, 80.56)
       };
       \addplot[fill=orange] coordinates {
           (GPT2, 83.8) (BERT, 86.2) (T5XL, 88.65) (T5XXL, 89.65)
       };
       \end{axis}
       \end{tikzpicture}
   \end{subfigure}
    \caption{Ablation studies. Top row: PathQuestion (PQ), middle row: MetaQA (MQ), and Bottom row: Multiple-choice datasets. From left to right: (1) Impact of subgraph selection threshold, (2) Effect of BFS exploration depth, (3) Comparison of prediction strategies (LLM/CLS/Hybrid), (4) Loss weighting influence ($\lambda_{cls}$), and (5) Performance across different language models (GPT2 through T5-XXL).}
    \label{fig:ablation_studies}
\end{figure*}

\subsection{Ablation Studies}
\label{sec:ablation}
We conduct extensive ablation studies to understand \system's performance and validate our design choices.

\noindent
\underline{Subgraph Selection Threshold.} The \textit{relation similarity threshold} ($\theta$) controlling subgraph extraction significantly influences model performance, with optimal values around 0.7 across all datasets. This finding contrasts with previous approaches like Q-KGR~\cite{zhang2024question} and GraphLLM~\cite{chai2023graphllm}, which rely on fixed-size neighborhood extraction. Our adaptive threshold enables more precise context selection, leading to 3-5\% accuracy improvements on complex queries.

\noindent
\underline{BFS Exploration Depth.} We observe that a \textit{BFS depth} of 3 consistently yields optimal results, achieving peak performance of 98\% on MetaQA and 90\% on multiple-choice tasks. The performance degradation at depth 4 indicates that controlled graph exploration is crucial for effective reasoning.

\noindent
\underline{LearningStrategy Comparison.} Our experiments compare three distinct entity prediction strategies: pure LLM reasoning (LLM), classification-based prediction (CLS), and our hybrid approach (HYB). While recent works like HiGPT~\cite{tang2024higpt} focus on enhancing LLM capabilities through specialized prompting, our hybrid strategy demonstrates that combining structured graph information with LLM reasoning yields superior results. This is particularly evident in complex scenarios like PQL-3h, where HYB achieves a 5\% accuracy gain over the pure LLM approach.

\noindent
\underline{Loss Balancing.} The classification \textit{loss weight }($\lambda_{cls}$) shows consistent optimal values between 0.5-0.7 across datasets, indicating an ideal balance between classification accuracy and LLM reasoning. Unlike methods that rely solely on either classification (e.g., QAGNN~\cite{WangRCB24}) or language modeling objectives, our balanced approach enables more robust learning. The sharp performance drop at $\lambda_{cls}=1.0$ validates the importance of maintaining LLM influence in the final prediction.

\noindent
\underline{LLM choice.} While larger LLMs generally yield better performance, with T5-XXL achieving the highest accuracy, \system\ demonstrates strong results even with smaller models. This contrasts with approaches like GraphToken~\cite{FatemiHP24} that show performance degradation with smaller LLMs. Our results suggest that \system's structured graph integration effectively leverages model capabilities, making it more practical for resource-constrained settings.
\section{Concluding Remarks and Future Work}
\label{sec:conclusions}
This paper introduced \system\ (The Graph Language), a novel framework that enables knowledge graphs to "speak" directly in the semantic space of large language models.  Our comprehensive analysis of \system\ demonstrates the effectiveness of bridging KGs and language models through semantic alignment rather than structural transformation. Through extensive experimentation, we have validated several key design principles that advance the state of the art in knowledge-intensive reasoning tasks. Our ablation studies reveal that controlled graph exploration (optimal at 3-hop depth) consistently outperforms both shallower and deeper alternatives across all datasets. The adaptive threshold mechanism for subgraph selection proves more effective than fixed neighborhood approaches, while our hybrid learning strategy demonstrates superior performance compared to pure LLM or classification methods. Notably, \system\ maintains robust performance across different model sizes, making it practical for various deployment scenarios. 

Building on these foundations, we identify several promising directions for future research: (i) extending \system\ to support self-evolving knowledge graphs where the system can predict new edges and update reasoning paths dynamically. This could enable continuous knowledge acquisition and refinement through interaction; (ii) developing GNNs that can effectively fuse visual and textual node features, expanding \system's applicability to rich multi-modal knowledge representations; (iii) investigating cross-domain transfer of reasoning patterns to leverage knowledge across different types of graphs and application domains.

\paragraph*{Supplemental Material Statement:} Source code for \system\ along with further dataset description and the complete pseudocode are available from GitHub\footnote{\url{https://github.com/giuseppepirro/gralan}}.
\bibliographystyle{splncs04}

\end{document}